%% file: main.tex
\documentclass[10pt, conference]{IEEEtran}
\IEEEoverridecommandlockouts
\usepackage{soul}
\usepackage{cite}
\usepackage{amsmath,amssymb,amsfonts}
\usepackage{algorithmic}
\usepackage{graphicx}
\usepackage{textcomp}
\usepackage{xcolor}
\usepackage{url}
\usepackage{booktabs}
\usepackage{caption}
\usepackage{subcaption}
\def\BibTeX{{\rm B\kern-.05em{\sc i\kern-.025em b}\kern-.08em
    T\kern-.1667em\lower.7ex\hbox{E}\kern-.125emX}}
    
\begin{document}

\title{The SERENADE project: Sensor-Based Explainable Detection of Cognitive Decline}

\author{
    \IEEEauthorblockN{
        Gabriele Civitarese\IEEEauthorrefmark{1},
        Michele Fiori\IEEEauthorrefmark{1},
        Andrea Arighi\IEEEauthorrefmark{3},
        Daniela Galimberti\IEEEauthorrefmark{2}\IEEEauthorrefmark{3},
        Graziana Florio\IEEEauthorrefmark{2}, 
        Claudio Bettini\IEEEauthorrefmark{1}
    }
    \IEEEauthorblockA{\IEEEauthorrefmark{1} EveryWare Lab, Dept. of Computer Science, University of Milan, Milan, Italy
    \\\{gabriele.civitarese, michele.fiori, daniela.galimberti, graziana.florio, claudio.bettini\}@unimi.com}
    \IEEEauthorblockA{\IEEEauthorrefmark{2} Dept. of Biomedical, Surgical and Dental Sciences, University of Milan, Milan, Italy
  }
    \IEEEauthorblockA{\IEEEauthorrefmark{3} Dept. of Neuroscience and Mental Health Area, Fondazione IRCCS Ca’ Granda Ospedale Maggiore Policlinico, Milan, Italy
      \\\{daniela.galimberti, andrea.arighi, lorem.ipsum\}@policlinico.mi.it}
}

\maketitle

\begin{abstract}
Mild Cognitive Impairment (MCI) affects $12$-$18\%$ of individuals over $60$. MCI patients exhibit cognitive dysfunctions without significant daily functional loss. While MCI may progress to dementia, predicting this transition remains a clinical challenge due to limited and unreliable indicators. %Detecting early cognitive decline is crucial
Behavioral changes, like in the execution of Activities of Daily Living (ADLs), can signal such progression. Sensorized smart homes and wearable devices offer an innovative solution for continuous, non-intrusive monitoring ADLs for MCI patients. However, current machine learning models for detecting behavioral changes lack transparency, hindering clinicians' trust. This paper introduces the SERENADE project, a European Union-funded initiative that aims to detect and explain behavioral changes associated with cognitive decline using explainable AI methods. SERENADE aims at collecting one year of data from 30 MCI patients living alone, leveraging AI to support clinical decision-making and offering a new approach to early dementia detection.
\end{abstract}


\section{Introduction}
\input{sections/intro}

\section{The SERENADE project}
\input{sections/serenade}

\section{SERENADE's Technological Setup}
\input{sections/setup}

\section{Deployment Experience}
\input{sections/experience}

\section{Data Analysis}
\input{sections/data-analysis}

\section{Conclusion and future work}
\input{sections/conclusions.tex}

%\section*{Acknowledgment}
%This work was supported in part by MUSA, SERICS, and FAIR projects under the NRRP MUR program funded by the EU-NGEU. Views and opinions expressed are those of the authors only and do not necessarily reflect those of the European Union or the Italian MUR. Neither the European Union nor the Italian MUR can be held responsible for them.

\bibliographystyle{ieeetr}
\bibliography{references}

\end{document}

%% file: sections/intro.tex
Approximately $12\%$ to $18\%$ of people over 60 years old live with Mild Cognitive Impairment (MCI), which is a pre-stage of dementia where the subject, even if presenting cognitive dysfunctions on one or multiple domains, still keeps their daily functionality. MCI may progress to dementia; however, such progression may not happen for each individual~\cite{gauthier2006mild}.

The evolution of MCI into dementia leads to a cognitive deficit that negatively impacts an individual's working, social, and relational capabilities. Indeed, the first dementia stages are often characterized by a progressive reduction of the functional autonomy of the subject in performing activities of daily living (ADLs). Significant changes in ADL habits (e.g., cooking, and personal hygiene activities) may serve as early indicators of cognitive decline~\cite{riboni2016smartfaber}, together with motor symptoms (e.g., slowing, tremor) and sleep disorders (e.g., insomnia, REM stage disorders)~\cite{pettersson2005motor,da2015sleep}.

The early detection of the transition from MCI to dementia is crucial. However, this is particularly challenging for clinicians since there are only a few unreliable indicators (e.g., patients' self-reporting) that can be used to predict the evolution. Moreover, due to the congestion in health systems, it is not possible to conduct frequent visits to continuously assess the development of MCI to dementia.

%Dunque, una regolare ed accurata valutazione clinica delle %funzioni cognitive è essenziale in tutti i pazienti con MCI. In particolare, è importante prendere in considerazione aree quali il linguaggio, le funzioni visuospaziali, la memoria e le funzioni executive.

%Ad oggi, un importante elemento per la valutazione indiretta dello stato funzionale di un paziente con MCI si basa sulle informazioni derivate dai parenti, che sono indicative delle necessità di cura del paziente stesso; tuttavia, tali informazioni sono spesso limitate dalla variabilità di assistenza fornita da parte del caregiver, e dalla percezione distorta delle reali abilità del paziente. Dunque, per ovviare a queste limitazioni, un’osservazione diretta delle performance nelle attività quotidiane del paziente sarebbe estremamente utile e permetterebbe una valutazione oggettiva della sua reale funzionalità. In tal senso, un ampio utilizzo di strumenti standardizzati permetterebbe di confrontare i pazienti, di osservare cambiamenti nel tempo e di determinare oggettivamente il bisogno di sostegno del paziente, così come l’impatto dei programmi terapeutici e riabilitativi impostati. A tal proposito, la possibilità di monitorare da remoto i cambiamenti comportamentali sopracitati apre le porte ad indicatori promettenti per la diagnosi precoce di demenza nei pazienti con MCI.

Smart homes and mobile/wearable devices represent a promising solution to continuously and unobtrusively monitor the behavioral habits of MCI subjects~\cite{lussier2018early,xie2019wearable}. %In addition, mobile/wearable devices (e.g., smartwatch, smartphone) make it possible to collect data from a high variety of domains, even outside the home environment (e.g., vital parameters, intensity of physical activities, trajectories). 
The intelligent analysis of data collected from such sensing devices represents a promising direction to early detect symptoms of cognitive decline, allowing clinicians to intervene promptly~\cite{fahad2021activity,gramkow2024digitized}.
%Sebbene l’utilizzo di telecamere e microfoni possa rappresentare un ulteriore soluzione per il monitoraggio continuo, questi dispositivi risultano eccessivamente intrusivi in termini di privacy.
%Le ADL rappresentano un aspetto fondamentale nella valutazione clinica dei pazienti con MCI e demenza: il loro riconoscimento automatico è essenziale per rilevare cambiamenti comportamentali indicativi di declino cognitivo. Infatti, diversi lavori di ricerca si sono focalizzati nell’analisi delle anomalie comportamentali nello svolgimento delle ADL per fornire ai clinici degli indicatori di declino cognitivo.
However, the majority of the approaches to detect activities and behavioral changes are based on opaque machine learning methods, whose decisions are often difficult to explain. This is a limit for clinicians who need to trust such automatic systems and interpret their results to support their diagnoses. 
%L’opacità dei sistemi di machine learning pone dei limiti anche ai tecnici che necessitano di comprendere al meglio come raffinare i modelli di riconoscimento e l’infrastruttura sensoristica. 

Recently, eXplainable Artificial Intelligence (XAI) has emerged as a powerful mechanism to interpret the decisions of machine learning models, providing human-readable explanations~\cite{arrotta2022dexar}. 
To the best of our knowledge, limited research efforts focused on XAI methods specifically for sensor-based cognitive decline detection~\cite{khodabandehloo2021healthxai}. 

In this paper, we introduce SERENADE~\footnote{\url{https://ecare.unimi.it/pilots/serenade}}, a three-year project pilot %funded by the European Union (NextGenerationEU) 
aimed at developing a technological platform capable of detecting and explaining behavioral changes %The goal of SERENADE is to detect behavioral changes 
across different domains (e.g., nutrition, sleep, mobility) to support clinicians in making therapeutic decisions for patients with MCI.  Specifically, SERENADE tackles the detection of anomalous changes in the execution of Activities of Daily Living (ADLs) and to correlate them with cognitive decline. During the SERENADE project, we plan to recruit $30$ subjects with MCI living alone, with the objective of collecting, for each subject, one year of sensor data from both smart home and wearable sensing devices. Behavioral change detection will be validated with periodic in-person visits carried out by clinicians.

While there have been many projects studying technical solutions to continuously monitor elderly subjects at home (e.g., CASAS~\cite{cook2012casas}, Gator SmartHome~\cite{helal2005gator}, %HOPE~\cite{hope}, 
Dem@care~\cite{karakostas2016care}, Hocare~\cite{hocare}, 
%Smart-Bear~\cite{kouris2020smart}, 
ENRICHME~\cite{cocsar2020enrichme}, %DOREMI~\cite{palumbo2016stigmergy}, 
and FrailSafe~\cite{papoutsi2018improving})%, and Frail~\cite{frail})
%they mainly focus on technological solutions rather than data analysis. On the other hand, 
the novelty of SERENADE relies on the development of XAI methods on unsupervised data to determine digital biomarkers of cognitive decline.

Most research studies in this area considered data collection in controlled lab settings~\cite{giannios2024semantic,grammatikopoulou2024assessing}, while in SERENADE we will collect data from the homes of the participants.
A closely related study was conducted in~\cite{palermo2023tihm}, which also released a public dataset (named TIHM). However, the data collection period for each participant was limited to less than two months. In contrast, SERENADE will gather data for each participant over a full year, also considering a broader range of unobtrusive sensing technologies to monitor several digital biomarkers that are of particular interest to clinicians.

Compared to existing works, SERENADE tackles two main challenges. The first is the unsupervised nature of our data. Indeed, existing XAI methods are typically based on fully supervised approaches, thus requiring vast amounts of data labeled with ground truth about activities and behaviors. However, collecting such information in real-world settings is not realistic. The second challenge is over-reliance: while we need to provide explanations that are intuitive and convincing for clinicians, a well-known risk is to also provide a sense of trust even when explanations are associated with mistakes of the AI algorithm~\cite{arrotta2022dexar}. 
%Indeed, explanations should provide indications about the possible mistakes performed by the data-driven model.

The contributions of this paper are the following:
\begin{itemize}
    \item We introduce the SERENADE project and its goals.
    \item We describe the technical solutions for behavioral monitoring and our  deployment experience.
    \item We present our preliminary results, outlining the research directions we intend to explore.
\end{itemize}

%% file: sections/serenade.tex
The SERENADE project has been approved by the ethical committee of Policlinico of Milan, which is one of the major hospitals in Milan, Italy. The research staff is composed of a) experts in sensor-based activity recognition and its applications in clinical settings and b) clinicians (i.e., neurologists and neuropsychologists) experts in cognitive decline, and c) technicians in charge of installing and monitoring the sensing infrastructures in the homes. 
%The clinicians in he indicators of cognitive decline that can be potentially be monitored with sensing devices, and to clinically validate the . Also, the patients and their caregivers will be involved, by providing feedback about the system and actively participating in improving it.

SERENADE aims to use eXplainable AI (XAI) approaches to identify, from unlabeled sensor data, abnormal behavioral changes that can suggest the evolution of MCI. %. that can also explain the reason behind the detected changes. 
Another objective is to assess the feasibility of a large-scale application of the remote monitoring system as an integral component of the visits of patients with MCI.

At the end of the project, we will evaluate three different aspects: i) the efficacy in detecting clinically relevant behavioral changes, ii) the quality of high-level explanations generated by the system, and iii) the acceptability of the sensing infrastructure from the patients' and caregivers' point of view.

\subsection{Overall architecture of SERENADE}

\begin{figure}[]
    \centering
    \includegraphics*[width=0.8\linewidth]{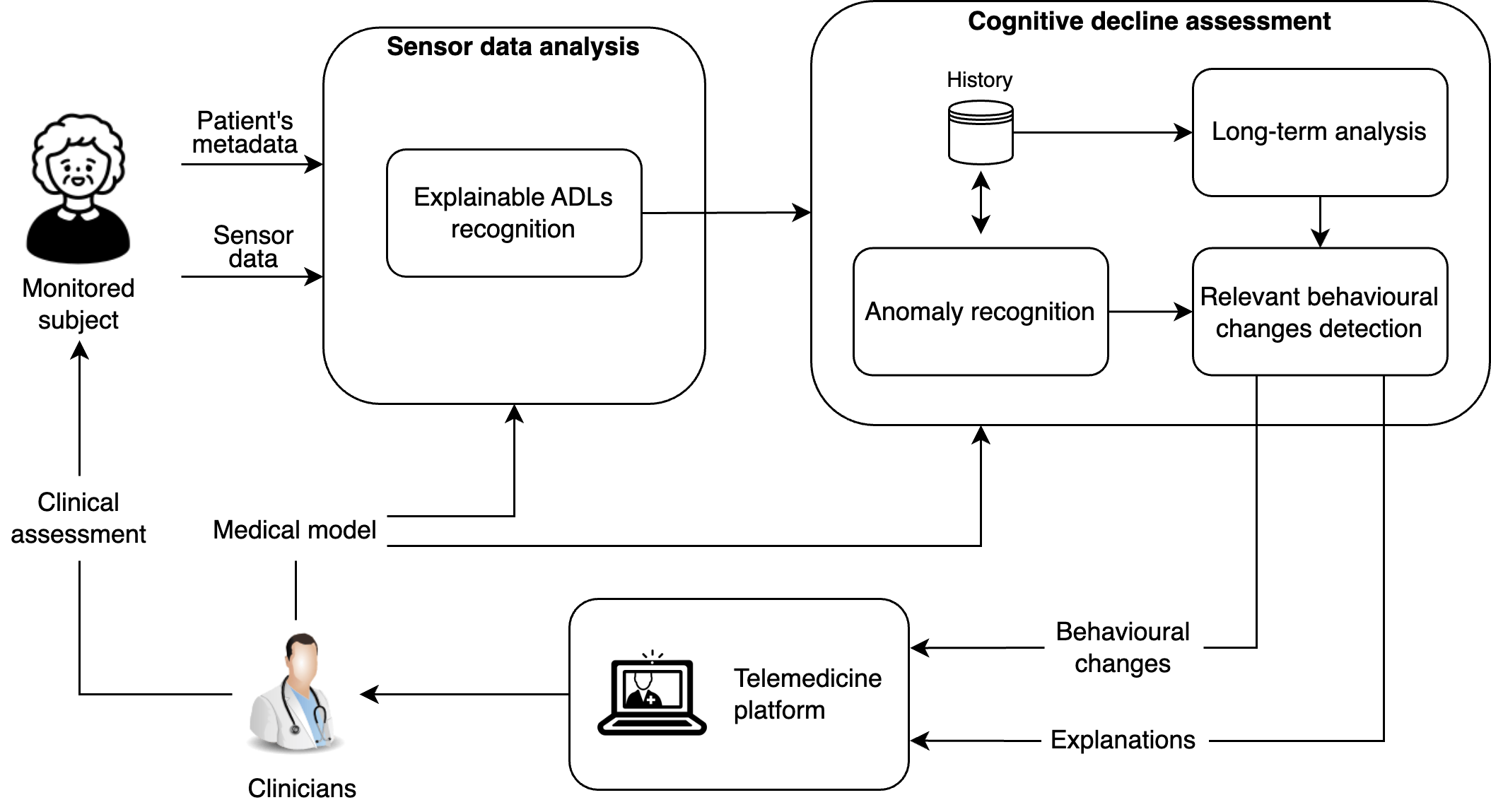}
    \caption{Overall SERENADE architecture}
    \label{fig:serenade}
\end{figure}

The overall SERENADE framework is depicted in Figure~\ref{fig:serenade}.
Sensor data is continuously collected in the home and processed by the \textsc{Sensor Data Analysis} module. This module is in charge of detecting high-level ADLs (e.g., cooking) adopting XAI approaches. This module also considers meta-information which are essential for recognizing ADLs, such as the home’s floorplan and sensor placement, which help identify activities accurately by understanding sensor data semantics. Additionally, knowing the caregiver's typical schedule is important, as its presence in the home significantly affects sensor data patterns. This meta-data is gathered by clinicians during visits and/or by technical teams during installation.

Detected ADLs are then analyzed by the \textsc{Cognitive Decline Assessment} module. This module leverages anomaly recognition and long-term analysis approaches to infer relevant long-term behavioral change possibly associated with cognitive decline.
Finally, detected behavioral changes and associated high-level explanations are transmitted to the \textsc{Telemedicine Platform}, where clinicians can inspect them. As we will explain later, clinicians provide periodic assessment of the monitored subjects to track the cognitive status and validate the system's output.

Note that, since our interest in this project is identifying digital markers of cognitive decline, the current framework does not include interventions or alerts, since they may implicitly alter the behavior of monitored subjects. 

\subsection{Cohorts}

SERENADE considers two cohorts of patients with MCI: the former where MCI is due to a neurodegenerative process that would likely lead to a progression in cognitive decline over time, and the latter where MCI is not due to neurodegeneration, where the disease development is often stable over time. The neurologists of the projects defined the inclusion criteria, that takes into account parameters like the age and the severity of MCI (according to neuroimaging and biological markers, carried out during the clinical diagnosis process).

%I pazienti verranno selezionati considerando: sesso, età, comorbidità, causa e severità dell’MCI, terapie farmacologiche in uso, presenza di un caregiver.
%La diagnosi di MCI implica la valutazione delle capacità decisionali del soggetto mediante test cognitivi, quali la Mini Mental Examination Scale (MMSE; metodo per la classificazione dello stato cognitivo dei pazienti per il medico Folstein MF, Folstein SE, McHugh PR. J Psychiatr Res. 1975 Nov; 12 (3): 189-98). Inoltre, l’indice di Barthel e il IADLs verranno utilizzati per misurare in modo oggettivo la dipendenza funzionale del paziente nello svolgere le sue attività quotidiane.
The target is to have $15$ subjects for each group, with the goal of testing the generalization capabilities of our methods. We are currently monitoring $18$ subjects ($7$ from the former group, $11$ for the latter). Each recruited subject had to sign a consent for data collection and analysis. 

\subsection{Behaviors of interest}

%The primary objective of SERENADE is to identify digital and unobtrusive biomarkers to monitor functional and symptomatic changes in MCI patients living alone. Such indicators should capture early the transition between MCI and dementia. 
After several meetings between the computer scientists, neurologists, and neuropsychologists participating in this project, we identified a list of clinically relevant behaviors that can be practically captured with sensing devices. Table~\ref{tab:behaviors} summarizes these behaviors, indicating some representative examples of clinically relevant changes, the associated ADLs, and the specific sensing devices that we use to monitor them.

\begin{table*}[t]
\tiny
\caption{Monitored behaviors}
\label{tab:behaviors}
\centering
\begin{tabular}{|l|l|l|l|}
\hline
\multicolumn{1}{|c|}{\textbf{Category}} & \multicolumn{1}{c|}{\textbf{Examples of Behavioral Changes}} & \multicolumn{1}{c|}{\textbf{Monitored Activities}} & \multicolumn{1}{c|}{\textbf{Devices}} \\ \hline

Nutrition &  \begin{tabular}[c]{@{}l@{}}Reduced or simplified cooking,\\ change of eating habits\end{tabular}   &     Meal Preparation, Eating   & \begin{tabular}[c]{@{}l@{}}Smart Plug (energy absorbed by\\ microwave and/or induction hob),\\ PIR/presence sensors,\\ door sensors on fridge \\
and food repositories,\\
temperature and humidity (stove)\end{tabular}                                   \\ \hline

Personal hygiene  & \begin{tabular}[c]{@{}l@{}} Reduced personal hygiene \end{tabular} &  Brushing teeth, using shower &    \begin{tabular}[c]{@{}l@{}}Temperature and humidity (shower),\\ Smart toothbrush,
PIR/presence sensors  \end{tabular} \\ \hline

Sleep  & \begin{tabular}[c]{@{}l@{}} Increased waking up during night,\\
increased movement while sleeping,\\ 
significant changes in sleep patterns\end{tabular} &  Different sleep phases, waking up events &    \begin{tabular}[c]{@{}l@{}}Smart mat sleep sensor,\\ PIR/presence sensors  \end{tabular} \\ \hline

Therapy  & Reduced adherence to medicine prescription &  Access to medicine repository &    \begin{tabular}[c]{@{}l@{}}Door sensor on medicine repository \end{tabular} \\ \hline

Mobility (at home)  & \begin{tabular}[c]{@{}l@{}}Increased sedentary behavior,\\emergence of wandering behavior,\\mobility patterns change \end{tabular} &  
 \begin{tabular}[c]{@{}l@{}}Presence and permanence in specific rooms,\\
movements and physical activities (steps)\end{tabular} &    \begin{tabular}[c]{@{}l@{}}PIR/presence sensors,\\Smartwatch  \end{tabular} \\ \hline

Mobility (outdoor)  & \begin{tabular}[c]{@{}l@{}}Decreased outdoor mobility,\\ 
significant changes in usual trajectories,\\
wandering behavior\end{tabular} &  
 \begin{tabular}[c]{@{}l@{}}Number of times the subject goes outside,\\
paths (duration, POIs, trajectories)\end{tabular} &    \begin{tabular}[c]{@{}l@{}}Smartwatch,\\Door sensor on entrance door  \end{tabular} \\ \hline

Cognition  & \begin{tabular}[c]{@{}l@{}}Refusing to take cognitive tests (apathy),\\
decreased cognitive tests performance\end{tabular} &  
 \begin{tabular}[c]{@{}l@{}}Cognitive tests' compliance and performance \end{tabular} &    \begin{tabular}[c]{@{}l@{}}Tablet (smart assistant with vocal interface) \end{tabular} \\ \hline

% Caregiver presence  & \begin{tabular}[c]{@{}l@{}}Occurrence of specific activities\\ only when the caregiver is at home,\\ increased caregiver visits\end{tabular} &  
% \begin{tabular}[c]{@{}l@{}}Caregiver presence\end{tabular} &    \begin{tabular}[c]{@{}l@{}}PIR sensors (to detect\\ multiple subjects at home),\\Door sensor on entrance door \end{tabular} \\ \hline

\end{tabular}
\end{table*}

%In the following, we summarize the sensing devices involved to capture the above-mentioned behaviors of interest.  Environmental sensors will capture the interactions of the subject with home infrastructure (i.e., door sensors, PIR sensors, plug sensors) as well as the temperature and humidity in specific home locations. A smart mat will be used to monitor sleep patterns. A smartwatch will keep track of physical activities Finally, a vocal interface (tablet) to periodically administer cognitive tests (designed by clinicians), in order to keep track of the cognitive status of the subject. 

\subsection{The patient's journey in SERENADE}
\label{sec:journey}

At the beginning of the study, each participant undergoes a thorough in-person medical evaluation, in order to assess the initial cognitive status. Following this visit, a member of the technical team performs a home inspection to document all the relevant technical details to outline the sensor placement. The inspection also serves to assess the feasibility of installing sensors and to verify that the required wireless signals have adequate coverage. Once this preparatory phase is complete, the sensing infrastructure is installed in the subject's home.
Throughout the project, the technical team may conduct additional inspections to address technical issues (e.g., resolving sensor malfunctions, or replacing devices with low battery levels).
Sensor data are continuously collected and stored in the telemedicine platform and analyzed by our algorithms. Prior follow-up visits, the medical staff reviews the detected behavioral changes and their associated explanations using a dedicated dashboard. In this project, the follow-up visits are planned after $6$ and $12$ months from the beginning, in order to track the patient's progress over time. 
The clinicians use the system's outputs to further support their diagnosis.

%The scientific findings of the SERENADE project will be validated by the international scientific community through publication and presentation at peer-reviewed conferences and journals. Technical and theoretical results will be directed to the computer science community, while experimental outcomes will be aimed at multidisciplinary research fields.

%The project is focused on basic technological research; it does not aim to develop new therapies nor collect patient-specific data that could directly inform individual treatment plans, as it does not involve the development or application of therapeutic strategies. The data analyzed in this project is not intended to have direct clinical relevance for ongoing treatment of individual patients.

%% file: sections/setup.tex
\label{sec:setup}
%In the following, we describe the technological setup adopted in SERENADE for data collection.

\subsection{Overall Sensing Infrastructure}
\begin{figure}[]
\centering
    \includegraphics[width=0.65\columnwidth]{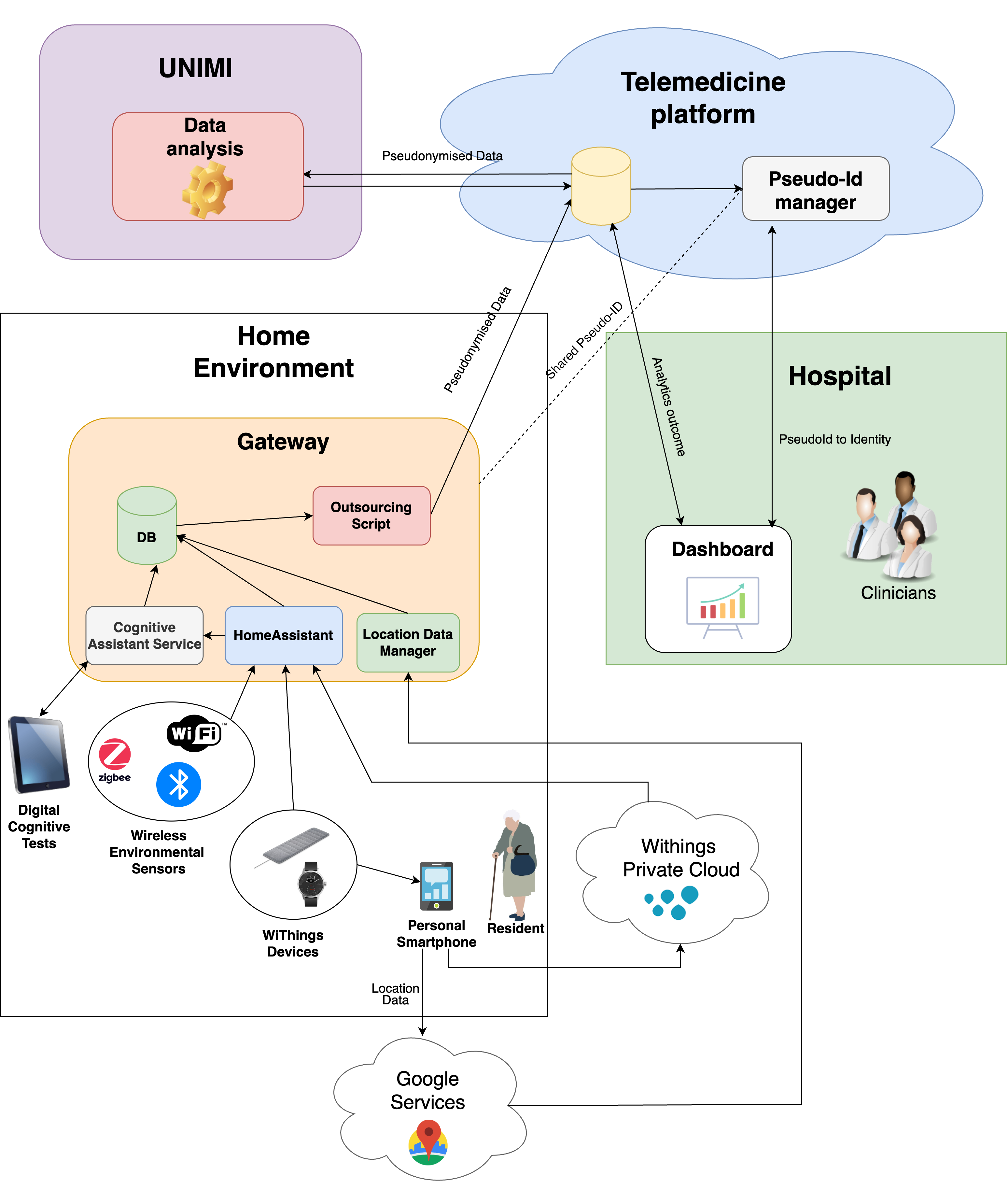}
    \caption{Our current technical setup}
    \label{fig:technicalsetup}
\end{figure}

Figure~\ref{fig:technicalsetup} shows the overall architecture of our data collection system.
The \textit{Home Environment} is the subject's home equipped with the sensing infrastructure. It continuously collects data, storing it on a local database. Periodically (i.e., every night), locally stored data are outsourced (using pseudonymization) to the telemedicine platform. The \textit{Telemedicine Platform} is a cloud service in a trusted domain, that securely stores data collected from the many installations, also providing a \textsc{Pseudo-id manager} module allowing clinicians to securely associate pseudonymized data with the subjects' identities.
\textit{UNIMI} is the entity currently in charge of running data analysis algorithms to detect behavioral changes. UNIMI works on pseudonymized data, so it can not easily re-identify the subjects. %Ideally, in a properly engineered system, the data analysis algorithms will run inside the \textsc{Telemedicine Platform}.
Finally, the \textit{Hospital} is the domain in which clinicians can inspect the data analysis results through a dedicated web dashboard.

\subsection{Sensing devices}

In the following, we describe the sensing devices adopted in the SERENADE project.

\subsubsection{Environmental Sensing Devices} unobtrusive off-the-shelf and relatively small objects monitoring the interaction of the subject with the home environment. They are installed in fixed positions. These devices communicate their readings through wireless networks (e.g., ZigBee or WiFi networks) to the home gateway.
Magnetic sensors detect the opening and closing of doors or drawers, such as kitchen pantries or fridges. A combination of motion (PIR) and presence (mmWave) sensors identify the presence of the subject in specific areas. Temperature and humidity sensors help infer events causing abrupt changes, such as a stove heating up or increased humidity during showers. Lastly, smart plugs monitor the usage of appliances like TVs, washing machines, or microwaves.

%Specifically, we consider the following sensors:

%\begin{itemize}
%    \item Magnetic Sensors: used to detect opening/closing of doors or drawers (e.g., kitchen pantry, fridge, house entrance). % use two pieces, a magnet, and a switch. When the magnet is close to the switch, it closes the circuit; when moved away, the circuit opens. 
%    \item Motion Sensors (PIRs): to detect motion in some specific area of the house, like the kitchen or the bathroom. They only trigger when detecting movement, requiring pre-processing to infer presence in certain home areas.
%    \item Presence Sensors: differently from PIRs, they continuously detect the presence of the subject, even if they remain stationary. We are using relatively new devices based on mmWave radars, with promising capabilities of subdividing a monitored room in different sub-locations.
%    \item Temperature and Humidity Sensors:  used to infer specific events that may cause abrupt changes in temperature/humidity values. For instance, the temperature over the stove suddenly increases when it is turned on. Similarly, the humidity near the shower significantly increases during showering events.
    %temperature sensors measure heat through changes in resistance or voltage, while humidity sensors detect moisture by measuring electrical changes in the air. 
%    \item Smart Plugs: used to detect the usage of specific appliances like TV, or washing machines, and microwave.
%\end{itemize}

\subsubsection{Sleep Analyzer}

we leverage the WiThings Sleep Analyzer as a thin smart mat deployed under the mattress. It monitors several sleep-related parameters, such as duration of overall sleep, heart rate, wake-up counts, duration of sleep phases (e.g., deep, REM), etcetera. 

\subsubsection{Smartwatch} 

we use the WiThings ScanWatch 2 as a smartwatch. Its design closely resembles that of a traditional watch, making it more acceptable and appealing to elderly subjects. This device collects several parameters related to physical activity, like the heart rate, the number of steps, etcetera.

\subsubsection{Location Data}

thanks to the subject's personal smartphone, we collect location data to track outdoor mobility through Google's Location API. The participants have the option to specifically consent for location data collection during the study.
%Further details on how such sensitive data is protected in terms of privacy will be explained later.

\subsubsection{Smart Toothbrush}

a toothbrush enabled with Bluetooth technology transmitting data about its usage, to monitor the patient's adherence to oral hygiene.

\subsection{Digital Cognitive Tests}

Our team of neuropsychologists developed a set of cognitive tests to periodically track subjects' cognitive function, assessing areas like memory, language, and orientation. These tests are delivered through an Android app on a tablet mounted on a wall. The system is voice-based, with auditory prompts for questions and the subject providing verbal responses. The app calculates scores automatically based on the responses.

Tests are scheduled weekly (based on the subjects' convenient times) and only begin when the system detects the subject's presence or movement near the tablet. When this happens, the app first asks the subject to confirm availability. If the subject agrees, the test starts; otherwise, it is rescheduled. After three missed attempts, the test is considered incomplete for the week. Note that a test is rescheduled even when there is no presence near the tablet after the scheduled time. %The schedule is set during installation, with the technician coordinating a convenient time. 

\subsection{Gateway}

The home gateway is a mini-computer 
%(in our setting, a Minisforum MiniPC) 
in charge of collecting and storing data from the many sensing devices. It leverages open-source software for domotics (HomeAssistant) to collect data from environmental sensors and the smart toothbrush. It also includes some additional modules to obtain data from digital cognitive tests and location data. The Gateway runs a local instance of InfluxDB to store collected data. Every night, collected data is outsourced to the Telemedicine Platform.

\subsection{Personal Data Protection}

To protect participants' identities during data analysis, we leverage pseudonymization. Sensor data transmitted by the home gateway to the telemedicine platform (i.e., a server in the hospital domain) is linked to a pseudonymous identifier instead of direct identifiers, making it nearly impossible for analysts to re-identify participants.

However, location data, which could potentially identify a subject even when pseudonymized, is encrypted before transmission. Only the location data analysis algorithm can decrypt this data to derive high-level insights without revealing sensitive details.

Clinicians, who need access to the analysis results for clinical purposes, are allowed to know participants' identities. This is managed through controlled access, with the identity re-association handled by the \textsc{Pseudo-id Manager} module on the Telemedicine Platform.

%% file: sections/experience.tex
We now describe the main challenges we are facing during deployment. %Currently, we deployed the system described in Section~\ref{sec:setup} in the homes of $18$ patients, and we are continuing to recruit. 

\subsection{Technical problems}

While we only adopt off-the-shelf sensing devices with the European Certificate of conformity, we implemented the whole data collection architecture, that is a working prototype but not yet engineered for large-scale deployments. As a consequence, frequent problems related to sensing failures and networking issues emerged. 
In order to minimize data disruptions, we implemented an alerting system providing real-time alerts (e.g., when the smart-home gateway is not reachable from outside, when a sensing device disconnected, etcetera). In addition, the alerting system also provides a daily report about missing data.

%Currently, our alerting system delivers information through a Telegram channel that is monitored by the research group. In the future, it will be implemented as an interactive web dashboard. 

When it is possible, the technical team attempts to solve the problems remotely (e.g., remotely accessing the server, or by asking the subject/caregiver to intervene when it is possible). However, several times it is necessary to physically visit the homes to address the issues.
%, especially for hardware problems or when the gateway is no longer remotely reachable. 

Sometimes, the technical problems are also related to improper handling of the devices by the participants, such as accidentally turning off the gateway or removing smart plugs. For instance, one patient discovered the sleep analyzer under the mattress but couldn't remember what it was, so they unplugged and dismissed it. 

\subsection{Acceptability}
%valutazione alla fine dello studio. risutati di questo seguiranno
Even though they could accurately capture activities, SERENADE does not use cameras or microphones since their use would be perceived as too obtrusive in home environments. Despite this, only $30\%$ of the individuals offered the opportunity to participate in the study accepted the invitation. This was due to skepticism about having sensors in the home or for a perceived violation of privacy. Other subjects didn't like the idea of having sensors and devices visible in their house both for aesthetic reasons and for confidentiality in letting any guests know they are being monitored. Unfortunately, the strict inclusion parameters required by the project (e.g., the subject should live alone, in the Milan area, and meeting the inclusion criteria defined by clinicians) also significantly reduced the number of potential candidates. %18 su 60

While at the end of our study we will carry out a thorough evaluation of acceptability, in the following we report our current experience. For instance, a patient explicitly asked us not to place the tablet in a position where it could be seen by potential visitors, as they felt embarrassed about it. Other participants expressed concerns about minor changes in their homes due to sensor deployments. Also, while we planned to collect outdoor location data, many subjects did not provide their consent, since it was perceived as too intrusive. Hence, we are collecting location data only for a small subset of the participating subjects.

%Another participant requested to install the tablet in a less conspicuous location, as they felt embarrassed about visitors noticing it.
Another important aspect is that, due to clinical reasons, our infrastructure does not provide immediate feedback to the patients themselves. Indeed, our current system studies the evolution of the digital indicators of cognitive decline without interfering with the patient's life, thus avoiding side effects in the behavior due to intervention. However, we observed that sometimes this generated skepticism (both from patients and caregivers) as the long-term advantages 
%are over a long period and
are not immediately perceived by the subjects. One patient withdrew from the study because their caregiver perceived no clear benefits in continuing and found the frequent technician visits to address technical issues burdensome. In SERENADE, we aim to mitigate this issue by ensuring that clinicians provide feedback to monitored subjects during follow-up visits, sharing insights and outcomes derived from the collected data.
 %add about cameras and the fact that monitoring requirements reduced acceptance rate. per primo

\subsection{Peculiarities of homes and subjects}

%parlare di iit che deve far mappa. serve ad analisti per interpretare gli stati. non identificante ma suff a far capire gli eventi rilevati
A challenging problem is that different homes have quite different characteristics in the number of rooms, their arrangement, places where it is possible to install sensors, monitorable appliances, etcetera. As a result, each installation is almost unique. In some homes, it may even happen that is not possible to install specific sensors. For instance, in one case, the shape of the entrance door did not allow the technicians to install the magnetic sensor. This heterogeneity makes the data analysis task very challenging.

During the first home inspection, the technical team also drafts a floorplan of the home to determine where to install the sensors. This floorplan, as well as the choice and the positioning of the sensors, is crucial to understanding the semantics of the events generated by the sensors, ensuring that data analysis is tailored for each home.

%% file: sections/data-analysis.tex
%data curation - come identificare attività - identificazione dei cambiamenti. per questa terza, stiamo inizialmente facendo analisi statistiche, più verso cosa ci orientiamo. anomalia diventa nuova normalità
%dashboard come strumento interattivo

\subsection{Data Curation and Analysis}

%A significant part of data analysis that we are currently carrying out is dedicated to cleaning the data obtained from the homes. Besides aiming at reducing the intrinsic noise of sensor data and applying typical data cleaning steps for time series, we also face the significant problem of missing data. Indeed, due to the technical problems that often occurs (as we described above) we sometimes have missing data from one, multiple sensors, or (even worse) all the sensors for a certain time period. 
Data cleaning is the most critical component of our analysis. We aim to reduce inherent noise in sensor measurements and tackle the significant issue of missing data. The technical problems described above frequently result in incomplete sensor readings, ranging from individual sensor failures to total sensor outages during specific time periods. Depending on the specific analysis and the size of each gap, it is possible to ignore missing data or to apply data imputation techniques.

%\subsection{Activity Recognition}

%In this phase of the project, we are still studying the activity recognition approaches that may work well on our unlabeled data. The most promising direction that we are evaluating is adopting self-supervised learning methods to learn a reliable representation from the collected unsupervised data. %This representation can be used to capture activity patterns and subjects' habits. 
While we are currently employing basic descriptive statistical methods for data analysis, we will explore activity recognition approaches for unlabeled data. The most promising direction we envision is adopting self-supervised learning methods to learn a reliable representation of sensor data to capture activity patterns and subjects' habits. 
Since we are working in an unlabeled setting, we will consider the adoption of approaches levarging common-sense knowledge to label the activity patterns like knowledge-based approaches~\cite{civitarese2019polaris} as well as solutions leveraging Large Language Models (LLMs)~\cite{civitarese2024large}. 

Among the many challenges, data segmentation plays a major role. We indeed hypothesize that fixed-time segmentation (as usually adopted in activity recognition) is not optimal in real-world settings, since different activities may have significantly different durations. We will explore dynamic segmentation strategies based on change point detection. Another significant challenge is that, unlike existing works, the set of possible activities performed by the subjects is not defined a priori.

%\subsection{Behavioral Changes Detection}

On the other hand, behavioral change detection methods typically involves defining an initial \textit{reference period} (e.g., the first few months of data collection) to serve as a baseline model, against which deviations are analyzed during a subsequent \textit{observation period} (e.g., the following months). Our focus is not on identifying individual anomalies or outliers within the reference period, but rather on detecting persistent abnormal behaviors that signify a shift to a sustained new activity patterns. As suggested by clinicians, a challenge is that some behavioral changes may simply occur due to seasonal variations (e.g., during winter a subject may reduce the time spent outside just because of the weather).  

Since we intend to leverage self-supervised learning for feature representation, an interesting direction we are planning is to detect behavioral changes by capturing concept drifts in the latent space. Our current plan is to adopt counterfactual approaches (i.e., determining which input features are the most likely cause of concept drift) in combination with LLMs  to generate explanations in natural language. 

%\subsection{Visualization and Diagnosis Support}

%We are currently implementing an interactive web dashboard that the clinicians will use to inspect the results of activity recognition and behavioral change detection methods. This dashboard will help the clinicians to support their diagnosis of cognitive decline during follow-up visit.
%We also plan to design a natural language interface for our dashboard that clinicians can use to query for specific clinical questions. 
Finally, we are implementing an interactive web dashboard for clinicians to examine activity recognition and behavioral change detection results. The dashboard will support clinicians in cognitive decline diagnosis during follow-up visits. Additionally, we plan to integrate a natural language interface to enable clinicians to perform targeted queries.
Similarly to the idea proposed in~\cite{gao2024leveraging}, we will leverage LLMs to automatically create the pipeline to generate the analysis and the plots required in the query.

\subsection{Preliminary Results on Behavioral Changes}

In the following, we show some preliminary results about potential behavioral changes we observed for a subject (we will refer to this subject as Alice). The clinicians have indicated that these initial insights show potential significance. %The clinicians confirmed us that these insights are potentially interesting. Currently, these results are obtained with basic descriptive statistics approaches.  

\begin{figure}[]
\centering
    \includegraphics[width=\columnwidth]{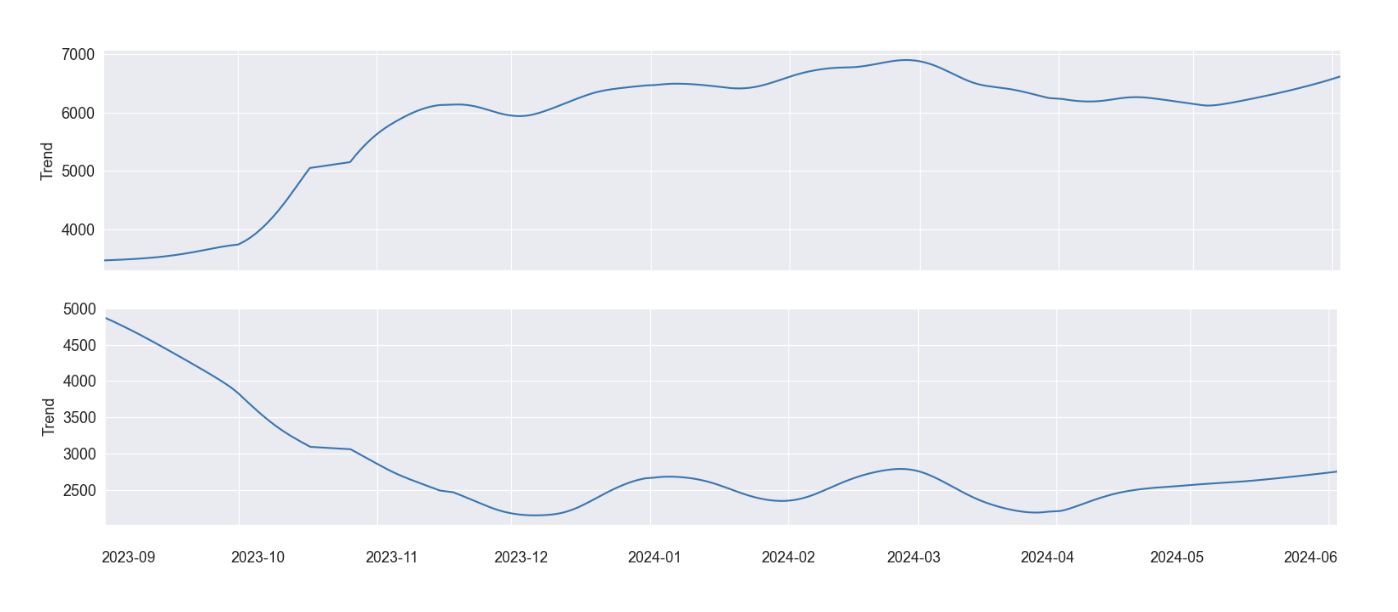}
    \caption{Sleep trends: REM sleep (top) increased, deep sleep (bottom) decreased.}
    \label{fig:sleep_trend}
\end{figure}

Among the relevant behaviors we monitor, we consider cooking habits. The temperature sensor deployed near the cooktop provides interesting insights about this behavior, since a peak is likely correlated with the subject cooking a hot meal (e.g., using the stove). For Alice, we observed a decrease in the number of temperature peaks during lunchtime. Under the clinicians suggestions, we also estimated the number of times the subject left the home (i.e., correlating the magnetic sensor on the door and the absence of activity) and we observed an increased number of going out during lunch time, potentially indicating that Alice may be eating out (or buying ready-to-eat food) more often. These findings highlight a potential shift in the subject's lunchtime habits. This trend is shown in Figure~\ref{fig:sleep_trend}.

\begin{figure}[]
\centering
    \includegraphics[width=0.8\columnwidth]{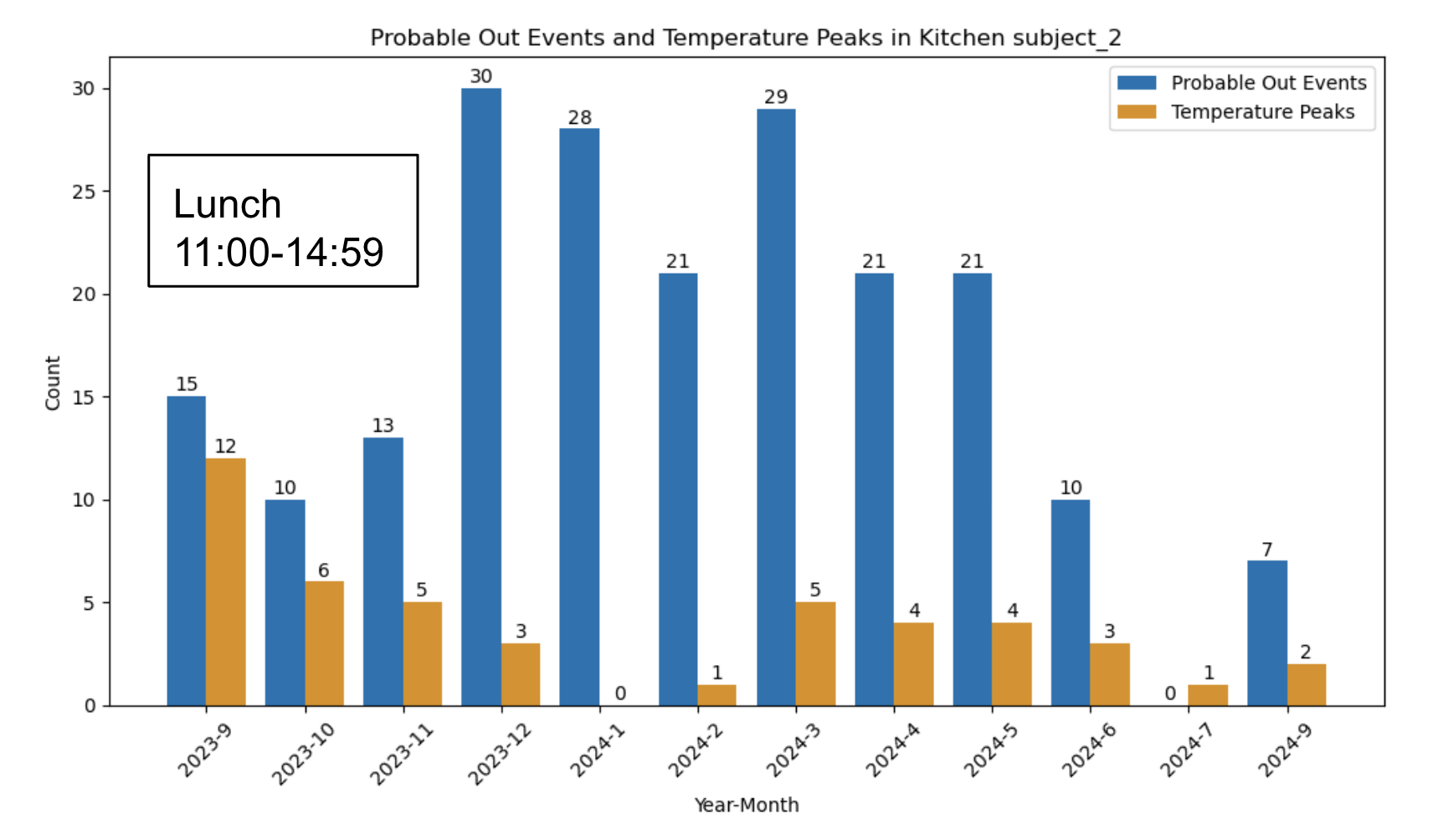}
    \caption{Nutrition: Temperature peaks over the suggests reduced cooking as outings during lunch increased.}
    \label{fig:peaks_vs_out}
\end{figure}

Interestingly, the sleep analyzer showed a potential behavioral change for Alice in the same observation period. Specifically, we observed a reduction in the duration of the deep sleep phase associated with an increase in REM sleep duration. The clinicians suggested that this pattern may be an indicator of disease evolution. This shift in sleep patterns is visible in Figure~\ref{fig:peaks_vs_out}.

%we can perform detailed data analysis by monitoring the duration of different sleep phases, including light sleep, deep sleep, and REM sleep, on a nightly basis. For one of the subjects under observation, the data reveals a notable trend starting in December 2023:  

%As explained before in the paper, these analysis are driven by the specific clinical questions of the clinicians. 

%In this phase is mostly based on feedback provided by clinicians in an iteration loop. Data analysis is guided by their specific clinical questions for the different monitored behaviors.

%In this preliminary phase, we use tools from statistical analysis to gain insight from the data.

%\subsection{Cooking events}

%\subsection{Sleep analysis}

%% file: sections/conclusions.tex
This paper outlined the objectives and experimental setup of the SERENADE project, which had been deployed in $18$ homes with ongoing recruitment efforts for additional participants. Although the sample size we are considering may be insufficient to draw clinically significant conclusions, the study will provide a solid technical foundation for large-scale deployments, addressing challenges and identifying key focus areas. From a medical perspective, the project contributes in exploring various indicators of cognitive decline, evaluating their diagnostic potential, and identifying the most promising ones. These findings may enable the development of a scalable, cost-effective, minimally intrusive, and easily maintainable infrastructure for future large-scale studies.

%We plan to use foundation models (SSL) and LLM

%We plan to we will have to complete the recruitment of all the $30$ subjects.

%Next steps in the project.

%% file: main.bbl
\begin{thebibliography}{10}

\bibitem{gauthier2006mild}
S.~Gauthier, B.~Reisberg, M.~Zaudig, R.~C. Petersen, K.~Ritchie, K.~Broich, S.~Belleville, H.~Brodaty, D.~Bennett, H.~Chertkow, {\em et~al.}, ``Mild cognitive impairment,'' {\em The lancet}, vol.~367, no.~9518, pp.~1262--1270, 2006.

\bibitem{riboni2016smartfaber}
D.~Riboni, C.~Bettini, G.~Civitarese, Z.~H. Janjua, and R.~Helaoui, ``Smartfaber: Recognizing fine-grained abnormal behaviors for early detection of mild cognitive impairment,'' {\em Artificial intelligence in medicine}, vol.~67, pp.~57--74, 2016.

\bibitem{pettersson2005motor}
A.~Pettersson, E.~Olsson, and L.-O. Wahlund, ``Motor function in subjects with mild cognitive impairment and early alzheimer’s disease,'' {\em Dementia and geriatric cognitive disorders}, vol.~19, no.~5-6, pp.~299--304, 2005.

\bibitem{da2015sleep}
R.~A. P.~C. da~Silva, ``Sleep disturbances and mild cognitive impairment: a review,'' {\em Sleep science}, vol.~8, no.~1, pp.~36--41, 2015.

\bibitem{lussier2018early}
M.~Lussier, M.~Lavoie, S.~Giroux, C.~Consel, M.~Guay, J.~Macoir, C.~Hudon, D.~Lorrain, L.~Talbot, F.~Langlois, {\em et~al.}, ``Early detection of mild cognitive impairment with in-home monitoring sensor technologies using functional measures: a systematic review,'' {\em IEEE journal of biomedical and health informatics}, vol.~23, no.~2, pp.~838--847, 2018.

\bibitem{xie2019wearable}
H.~Xie, Y.~Wang, S.~Tao, S.~Huang, C.~Zhang, and Z.~Lv, ``Wearable sensor-based daily life walking assessment of gait for distinguishing individuals with amnestic mild cognitive impairment,'' {\em Frontiers in aging neuroscience}, vol.~11, p.~285, 2019.

\bibitem{fahad2021activity}
L.~G. Fahad and S.~F. Tahir, ``Activity recognition and anomaly detection in smart homes,'' {\em Neurocomputing}, vol.~423, pp.~362--372, 2021.

\bibitem{gramkow2024digitized}
M.~H. Gramkow, G.~Waldemar, and K.~S. Frederiksen, ``The digitized memory clinic,'' {\em Nature Reviews Neurology}, pp.~1--9, 2024.

\bibitem{arrotta2022dexar}
L.~Arrotta, G.~Civitarese, and C.~Bettini, ``Dexar: Deep explainable sensor-based activity recognition in smart-home environments,'' {\em Proceedings of the ACM on Interactive, Mobile, Wearable and Ubiquitous Technologies}, vol.~6, no.~1, pp.~1--30, 2022.

\bibitem{khodabandehloo2021healthxai}
E.~Khodabandehloo, D.~Riboni, and A.~Alimohammadi, ``Healthxai: Collaborative and explainable ai for supporting early diagnosis of cognitive decline,'' {\em Future Generation Computer Systems}, vol.~116, pp.~168--189, 2021.

\bibitem{cook2012casas}
D.~J. Cook, A.~S. Crandall, B.~L. Thomas, and N.~C. Krishnan, ``Casas: A smart home in a box,'' {\em Computer}, vol.~46, no.~7, pp.~62--69, 2012.

\bibitem{helal2005gator}
S.~Helal, W.~Mann, H.~El-Zabadani, J.~King, Y.~Kaddoura, and E.~Jansen, ``The gator tech smart house: A programmable pervasive space,'' {\em Computer}, vol.~38, no.~3, pp.~50--60, 2005.

\bibitem{karakostas2016care}
A.~Karakostas, A.~Briassouli, K.~Avgerinakis, I.~Kompatsiaris, and M.~Tsolaki, ``The dem@ care experiments and datasets: a technical report,'' {\em arXiv preprint arXiv:1701.01142}, 2016.

\bibitem{hocare}
https://projects2014-2020.interregeurope.eu/hocare.

\bibitem{cocsar2020enrichme}
S.~Co{\c{s}}ar, M.~Fernandez-Carmona, R.~Agrigoroaie, J.~Pages, F.~Ferland, F.~Zhao, S.~Yue, N.~Bellotto, and A.~Tapus, ``Enrichme: Perception and interaction of an assistive robot for the elderly at home,'' {\em International Journal of Social Robotics}, vol.~12, pp.~779--805, 2020.

\bibitem{papoutsi2018improving}
C.~Papoutsi, A.~Poots, J.~Clements, Z.~Wyrko, N.~Offord, and J.~E. Reed, ``Improving patient safety for older people in acute admissions: implementation of the frailsafe checklist in 12 hospitals across the uk,'' {\em Age and ageing}, vol.~47, no.~2, pp.~311--317, 2018.

\bibitem{giannios2024semantic}
G.~Giannios, L.~Mpaltadoros, V.~Alepopoulos, M.~Grammatikopoulou, T.~G. Stavropoulos, S.~Nikolopoulos, I.~Lazarou, M.~Tsolaki, and I.~Kompatsiaris, ``A semantic framework to detect problems in activities of daily living monitored through smart home sensors,'' {\em Sensors}, vol.~24, no.~4, p.~1107, 2024.

\bibitem{grammatikopoulou2024assessing}
M.~Grammatikopoulou, I.~Lazarou, V.~Alepopoulos, L.~Mpaltadoros, V.~P. Oikonomou, T.~G. Stavropoulos, S.~Nikolopoulos, I.~Kompatsiaris, and M.~Tsolaki, ``Assessing the cognitive decline of people in the spectrum of ad by monitoring their activities of daily living in an iot-enabled smart home environment: a cross-sectional pilot study,'' {\em Frontiers in Aging Neuroscience}, vol.~16, p.~1375131, 2024.

\bibitem{palermo2023tihm}
F.~Palermo, Y.~Chen, A.~Capstick, N.~Fletcher-Loyd, C.~Walsh, S.~Kouchaki, J.~True, O.~Balazikova, E.~Soreq, G.~Scott, {\em et~al.}, ``Tihm: An open dataset for remote healthcare monitoring in dementia,'' {\em Scientific data}, vol.~10, no.~1, p.~606, 2023.

\bibitem{civitarese2019polaris}
G.~Civitarese, T.~Sztyler, D.~Riboni, C.~Bettini, and H.~Stuckenschmidt, ``Polaris: Probabilistic and ontological activity recognition in smart-homes,'' {\em IEEE T KNOWL DATA EN}, vol.~33, no.~1, pp.~209--223, 2019.

\bibitem{civitarese2024large}
G.~Civitarese, M.~Fiori, P.~Choudhary, and C.~Bettini, ``Large language models are zero-shot recognizers for activities of daily living,'' {\em arXiv preprint arXiv:2407.01238}, 2024.

\bibitem{gao2024leveraging}
N.~Gao, Z.~Yu, Y.~Xu, C.~Yu, Y.~Wang, F.~D. Salim, and Y.~Shi, ``Leveraging large language models for generating mobile sensing strategies in human behavior modeling,'' in {\em Companion of the 2024 on ACM International Joint Conference on Pervasive and Ubiquitous Computing}, pp.~729--735, 2024.

\end{thebibliography}
